\documentclass[sigconf,nonacm]{aamas}

\usepackage{balance} 
\usepackage[switch]{lineno}

\usepackage{multirow}
\usepackage{changepage}
\usepackage{hhline}

\newcommand{\triplebackticks}{\`{}\`{}\`{}}

\acmSubmissionID{853}

\title[LLM-augmented Preference Learning]{LLM-augmented Preference Learning from Natural Language}

\author{Inwon Kang$^1$, Sikai Ruan$^1$, Tyler Ho$^1$, Jui-Chien Lin$^1$, Farhad Mohsin$^2$,\\ Oshani Seneviratne$^1$, Lirong Xia$^1$}
\affiliation{
  \institution{$^1$Rensselaer Polytechnic Institute, Troy, NY, USA\\ $^2$College of the Holy Cross, Worcester, MA, USA}
  \city{}
  \country{}
  }
\email{}

\begin{abstract}
Finding preferences expressed in natural language is an important but challenging task.
State-of-the-art(SotA) methods leverage transformer-based models such as BERT, RoBERTa, etc. and graph neural architectures such as graph attention networks.
Since Large Language Models (LLMs) are equipped to deal with larger context lengths and have much larger model sizes than the transformer-based model, we investigate their ability to classify comparative text directly.
This work aims to serve as a first step towards using LLMs for the CPC task.
We design and conduct a set of experiments that format the classification task into an input prompt for the LLM and a methodology to get a fixed-format response that can be automatically evaluated.
Comparing performances with existing methods, we see that pre-trained LLMs are able to outperform the previous SotA models with no fine-tuning involved.
Our results show that the LLMs can consistently outperform the SotA when the target text is large -- i.e. composed of multiple sentences --, and are still comparable to the SotA performance in shorter text.
We also find that few-shot learning yields better performance than zero-shot learning.

\end{abstract}


\newcommand{\BibTeX}{\rm B\kern-.05em{\sc i\kern-.025em b}\kern-.08em\TeX}

\begin{document}


\pagestyle{fancy}
\fancyhead{}


\maketitle 


\section{Introduction}

Group decision making is an important task in multi-agent systems, in which a group of agents aim to make a collective decision based on their preferences. With the advent of artificial intelligence and machine learning techniques, finding better ways of making group decisions efficiently and accurate becomes an important problem~\cite{xia2022group}. 
In particular, state-of-the-art natural language processing (NLP) techniques have been used to improve preference elicitation, learning, and aggregation. For example, learning preferences from group discussions such as those in forums or chat rooms can provide an unobtrusive way of learning preferences and making a group decision. Learning preferences from text will also help in making better decisions when there are a large number of alternatives or when there is uncertainty about the preferences.

\citet{mohsin2021making} proposed a framework for making group decisions from natural language and created a dataset for texts with preferences. However, making group decisions turned out to be a difficult problem because predicting individual preferences is a difficult task by itself. 
Predicting preferences in text can be as simple as using simple grammatical rules. For example, look for expressions such as: ``better than'': the sentence ``Tea is better than coffee'' expresses a straightforward preference towards tea. 
However, human language is capable of much more nuanced expressions as well. 
Consider the sentence ``I used to think tea is better than coffee. However, that was a while ago.''.
A glance at the sentence may suggest that the author prefers tea over coffee but the sentence describes the opposite of the preference. 
While more than this example is needed to confuse most English speakers, such nuances in natural language add to the challenges of preference classification with machine learning (ML) models.
\citet{panchenko-etal-2019-categorizing} and~\citet{ma2020edgat-compsent} proposed methods predicting the preference between two alternatives from individual pieces of texts using modern machine learning techniques such as transformer-based word embeddings~\cite{pennington2014glove,devlin2019bert} and graph neural networks. However, the benchmark dataset~\cite{panchenko-etal-2018-building} considered only simple texts consisting of single sentences, and all texts contained mention of both alternatives. \citet{haque2022pixie} and \citet{mohsin2023cpc} proposed methods to improve these performances for more complex but realistic text that included implicit mention of alternatives and multi-sentence texts. However, unlike tasks like sentiment analysis and stance detection, where these NLP techniques achieved high accuracy, predicting preferences is a more difficult task with relatively low accuracy levels.

Large language models (LLMs) and foundation models have brought in a new wave of improvements in artificial intelligence systems. There is potential to use these LLMs in various multi-agent scenarios for different problems, including negotiation, delegation and making group decisions. Quite recently, Meta released a new feature in their chatting app, that allows an LLM-powered chatbot to help make group decisions\footnote{\url{https://about.fb.com/news/2023/09/introducing-ai-powered-assistants-characters-and-creative-tools/}}. However, preference aggregation, deliberation, etc. depends on the language model's ability to identify preferences expressed in the text. The question is: 
\begin{center}
    \bf Can LLMs identify comparative preferences in texts?
\end{center}{}

\noindent {\bf Our Contributions}
We investigate popular LLM's ability to identify and predict preferences in text in this work. For this, we use two benchmark datasets: First, the Compsent-19 dataset~\cite{panchenko-etal-2019-categorizing} which has single sentence texts with mentions of both alternatives, and second, the College Confidential dataset, which contains more complex texts. 
In particular, we experiment with two versions of Meta's LLaMa-2 model~\cite{touvronLLaMAOpenEfficient2023a}, the 13B-parameter version and the 4-bit quantized 70B-parameter version. We also considered OpenAI's popular GPT-3.5-Turbo model and GPT-4 model~\cite{openaiGPT4TechnicalReport2023}. 
We design and experiment with different prompts that ask the LLMs to predict the preference expressed in the text to determine what type of prompt results in the best prediction performance. Our key findings are the following.
\begin{center}
    \textbf{Key finding 1: LLMs can outperform previous state-of-the-art models.}
\end{center}{}
\noindent 
Results from our experiments show that the LLMs are able to outperform previous state-of-the-art models by just using few examples.
The best performance comes from the largest model (GPT-4), but the results show that even the smaller models (LLaMa-2-70B) are still able to outperform the state-of-the-art models. 
In the College Confidential tasks where the text is longer and has a more complex grammatical structure, almost all instances of the prompt/example configuration on both GPT-4 and LLaMa-2-70B outperform the state-of-the-art performance. 

\begin{center}
    \textbf{Key finding 2: Few-shot learning outperforms zero-shot learning.}
\end{center}{}
\noindent  
We use the zero-shot and few-shot prompts in combination with different styles of instruction to arrive at this conclusion.
Our prompt design was motivated by work in LLM literature that indicated that LLMs can behave as zero-shot~\cite{kojimaLargeLanguageModels2022,weiFINETUNEDLANGUAGEMODELS2022a} or few-shot~\cite{brownLanguageModelsAre2020} learners.
Our results show that few-shot learning almost always outperforms zero-shot learning in most models. 
We also find that smaller models may struggle with handling both the few-shot examples and longer prompts. 
But if the model is powerful enough -- e.g. GPT-4 --, providing detailed instruction with real-life examples yields the best performance.
\\

\begin{center}
    \textbf{Key finding 3: LLMs have superior performance in large and complex texts.}
\end{center}{}
\noindent  
The text in College Confidential dataset contains multiple sentences, complex grammatical structure, and pronouns; many mention the same entity more than once. All of these make College Confidential a challenging dataset. Fortunately, we found that LLMs have the ability to handle large and complex context, especially the LLMs with large number of parameters.
The size of the models proved to have some effect on the performance.
In the case of the smallest model (LLaMa-2-13B), getting it to predict preferences in a well-formatted way was difficult, and when it did predict preferences, it was mostly incorrect.
On the one hand, for the LLaMa-2-70B model -- which is smaller than GPT-3.5-Turbo--, we developed a retry prompt process that is an iterative process (Figure~\ref{fig:retry-prompt}) that manages to get well-formatted predictions from the LLM.
On the other hand, we did not need this iterative process for OpenAI's GPT-3.5-Turbo and GPT-4 as it produced perfectly formatted responses which allowed for automatic evaluation.
Interestingly, both LLaMa-2-70B and GPT-4 significantly outperformed state-of-the-art methods using transformer-based embeddings and graph neural networks for both simple and complex texts (the Compsent-19 and College Confidential datasets).
We find that both LLMs perform similarly for Compsent-19 (the single sentence benchmark). And for College Confidential, GPT-4 performs better than LLaMa-2-70B.

While these findings are not surprising given the versatility of LLMs, we note that these performances were a result of non-trivial deliberation.
Our methodology allows the LLM to be used without human intervention.
Because the LLMs do not output in a fixed format -- only restricted to English language --, we design a method to coerce the model to output its responses in a fixed format which can then be integrated into a larger pipeline in an automated way. 

This work aims to serve as the first step towards integrating LLMs in the CPC task.
While the previous state-of-the-art models have their advantage in some aspects, our results show that LLMs are able to effectively handle the CPC task with proper prompting techniques and are able to outperform them, even without fine-tuning.


\subsection{Related Works}
\noindent\textbf{Preference learning from text.}
A common approach for learning preferences from agents when wanting to make a group decision is preference elicitation. This involves asking interactive questions 
to efficiently elicit agents’ preferential information that is sufficient for making a group decision (e.g., in ~\cite{Boutilier02:POMDP,Conitzer02:Elicitation,Mandal2020:Optimal,Zhao2018:A-Cost-Effective}). \citet{xia2019learning} gives a good exposition to different preference learning methods.  ~\citet{mohsin2021making} proposed an unobtrusive framework of learning preferences from text (such as in chat rooms or forums) to make group decisions.

\citet{panchenko-etal-2019-categorizing} built the Compsent-19 dataset with the specific goal of categorizing comparative sentences and finding expressed preferences. 
\citet{panchenko-etal-2019-categorizing} used pre-trained sentence embeddings~\citep{conneau-EtAl:2017:EMNLP2017,pennington2014glove}, among other features, to train for the classification problem.
Since then, other machine learning techniques have been used for predicting preference.
Transformer models~\citep{vaswani2017attention}, in particular, have been useful tool in NLP because of their ability to learn large dimensions of data without overfitting. 
This has led to newer embeddings, both at token and sentence levels~\cite{devlin2019bert,liu2019roberta,gao2021simcse}.
Previous works have used text embeddings and graph neural networks like GAT\citep{velickovic2018graph} for preference detection/tagging. These models are effective with textual and graph-structured data.
Recent works in preference learning from text~\citep{huangSyntaxAwareAspectLevel2019,ma2020edgat-compsent,li2021powering} have made use of dependency graphs and graph neural network methods. ~\citet{li2021powering} additionally used the knowledge transfer technique from the related sentiment analysis task. But all of these methods were tested on the Compsent-19 benchmark~\citep{panchenko-etal-2019-categorizing} and thus dealt with single sentences with explicit mention of alternatives.
On the other hand, \citet{mohsin2021making} and~\citet{haque2022pixie} introduced new datasets which dealt with implicit preferences, where both entities might not be mentioned. The College Confidential dataset~\cite{mohsin2021making} in particular contained multi-sentence texts in a discussion setting. We work with both the simpler Compsent-19 dataset and the more complex College Confidential dataset.

\noindent\textbf{Large Language Models.} The introduction of LLMs and foundation models~\citep{chowdheryPaLMScalingLanguage2022,radfordImprovingLanguageUnderstanding2018,touvronLLaMAOpenEfficient2023a, openaiGPT4TechnicalReport2023} such as ChatGPT, LLaMa and Bard have brought forth a new paradigm in many domains of machine learning. In particular, ChatGPT~\citep{openaiIntroducingChatGPT2022} has demonstrated an impressive ability to perform various tasks while displaying human-like qualities through text. While these models mainly operate in natural language domain, past works have found that they can be well suited for other tasks, such as classification or regression~\citep{hegselmannTabLLMFewshotClassification2023, brownLanguageModelsAre2020, weiFINETUNEDLANGUAGEMODELS2022a}. However, the input to an LLM still needs to be encoded into human language, and past works have experimented with different ways of converting the tasks into natural language.

~\citet{brownLanguageModelsAre2020} demonstrate that few-shot learning can be utilized to enhance the performance of LLMs. In few-shot learning, the user includes correct examples of the given task to help the model's understanding of the task. ~\citet{weiFINETUNEDLANGUAGEMODELS2022a} show that LLMs can be powerful zero-shot learners when fine-tuning is applied to the pre-trained weights. 
~\citet{weiChainofThoughtPromptingElicits2022} use chain-of-thought prompting to guide the LLM towards making a step-by-step decision and show that this style of prompting can achieve better performance in tasks such as arithmetic or reasoning-based tasks.
In addition, ~\citet{kojimaLargeLanguageModels2022} show that LLMs can have \textit{decent} performance in zero-shot settings by using the chain-of-thought prompting process. ~\citet{schickExploitingClozeQuestionsFewShot2021} introduce Pattern Exploiting Training (PET), in which the prompt follows some pattern to elicit a higher quality of response from the LLM.

LLMs have proven to be proficient in non-textual settings as well.
~\citet{hegselmannTabLLMFewshotClassification2023} explore using LLMs to predict tabular data in a few-shot setting. The benchmark results show that LLMs are able to outperform previous state-of-the-art models, and that even zero-shot learning can achieve nontrivial performance in many instances. 


\section{Preliminaries}

        

\subsection{Task Description}
\label{cpc-description}

Given a text $t$, and a two alternatives $A$ and $B$, the goal is to predict the preference relation between $A$ and $B$. Ideally, there can be four possible cases: $A$ is preferred to $B$ ($A \succ B$), $B$ is preferred to $A$ ($A \prec B$), both are equally preferred ($A = B$), and there is no preference relation between the two alternatives ({\it N/A}).


This task has sometimes been called comparative preference classification (CPC). We consider two preference datasets in this work: College Confidential~\citep{mohsin2021making} and Compsent-19~\citep{panchenko-etal-2018-building}. The College Confidential dataset~\cite{mohsin2021making} contains comments from a college admission forum.
The authors search for discussion threads where the original poster asks for opinion comparing multiple colleges and collect the following posts to build the dataset. 
The discussion threads discuss more than two colleges in some cases. 
The multi-way comparisons are divided into pairwise comparisons.
The resulting dataset consists of 2964 pairwise comparison instances, with 4 classes -- \textit{No Preference}, $A \succ B$, $A \prec B$, $A = B$.

Compsent-19~\citep{panchenko-etal-2018-building} is a binary-comparative dataset that contains a single comparative sentence between two alternatives. 
The alternatives are picked from various domains such as computer science concepts -- programming languages, hardware devices --, or brands. 
The authors query for sentences that contain mentions of both alternatives from the Common Crawl dataset and present a final dataset of 7,199 sentences with 217 unique pairs of alternatives and 3 classes --  $A \succ B$, $A \prec B$, \textit{N/A}.

Because of the difference in the class representation of the two datasets, we only consider the 3 class cases in this work -- $A \succ B$, $A \prec B$, \textit{N/A} --, where \textit{N/A} refers to both \textit{No Preference} and $A = B$. The distribution of labels, along with the average text size for both datasets, is given in Table~\ref{tab:stat_data}. The average text size is given in token numbers, where tokens are building blocks of sentences in NLP. Words, punctuations, and parts of words can all be individual tokens. From this, we see that the College Confidential dataset has text that is, on average, more than four times longer than those in Compsent-19. Tables~\ref{tab:colcon-discussion} and~\ref{tab:compsent-discussion} show a few example texts, along with labels, from each dataset.

\begin{table*}[ht]
\begin{tabular}{|l|llll|r|}
\hline
\multirow{2}{*}{} & \multicolumn{4}{l|}{Label Distribution} & \multirow{2}{*}{Average Token Length} \\ \cline{2-5}
                     & \multicolumn{1}{l|}{A $\succ$ B} & \multicolumn{1}{l|}{A $\prec$ B} & \multicolumn{1}{l|}{N/A}  & Total &        \\ \hline
College Confidential & \multicolumn{1}{l|}{598}              & \multicolumn{1}{l|}{544}           & \multicolumn{1}{l|}{1822} & 2964  & 116.12 \\ \hline
Compsent-19          & \multicolumn{1}{l|}{1364}             & \multicolumn{1}{l|}{593}           & \multicolumn{1}{l|}{5242} & 7199  & 26.94  \\ \hline
\end{tabular}
\caption{Statistics for the datasets}
\label{tab:stat_data}
\end{table*}

\begin{table}[ht]
	\centering
	\small
	\begin{tabular}{p{0.7\linewidth}|l}
		\textbf{Sentence} & \textbf{Label}       \\
		\midrule
		If Duke is more expensive then go to UCB. Your parents will thank you for saving them money by going there. And you will have more access to jobs in Calif when you graduate. To me its a no brainer.                       & $C \succ D$ \\\hline
		Daughter graduating from Cal next month. full disclosure. To me, Cal is a no brainer.      & $C \succ D$ \\\hline
		Both are really good schools. You cannot   make a mistake going to either place.              & N/A           
	\end{tabular}
	\caption{Examples from the College Confidential dataset}
	\label{tab:colcon-discussion}
\end{table}

\begin{table}[ht]
	\centering
	\small
	\begin{tabular}{p{0.7\linewidth}|l}
		\textbf{Sentence} & \textbf{Label}       \\
		\midrule
		Golf is easier to pick up than baseball.                       & $g \succ b$ \\\hline
		I’m considering learning Python and more PHP if any of those would be better.      & N/A           
	\end{tabular}
	\caption{Examples from the Compsent-19 dataset}
	\label{tab:compsent-discussion}
\end{table}

\subsection{NLP and ML Terminology}
In this section, we discuss the various NLP techniques that we consider in this work.

\noindent\textbf{Large language models (LLM)} usually refer to pre-trained models that use the attention mechanism. What distinguishes an LLM from a regular language model is the \textit{large} amount of data used for pre-training and the number of parameters in the architecture, which are in the order of billions. 
We will only refer to large-scale transformer-based~\citep{vaswani2017attention} models that are pre-trained on massive corpuses, such as GPT~\citep{openaiIntroducingChatGPT2022, openaiGPT4TechnicalReport2023} and LLaMa~\citep{touvronLLaMAOpenEfficient2023a} as LLMs to avoid any confusion.
In this work, we consider OpenAI's GPT-4~\citep{openaiGPT4TechnicalReport2023} as the state-of-the-art LLM and use a fine-tuned version of Meta's LLaMa-2~\citep{touvronLLaMAOpenEfficient2023a}'s 70B model\footnote{This was finetuned by Upstage for instructions. \url{https://huggingface.co/TheBloke/Upstage-Llama-2-70B-instruct-v2-GPTQ}} and the original 13B chat model\footnote{\url{https://huggingface.co/TheBloke/Llama-2-13B-chat-GGML}} as the open-source alternative. 
Due to the memory constraints, we use the 4-bit quantized version of the 70B model\footnote{We use a machine with a single A6000 with 48G VRAM}.
We will refer to these models as LLaMa-2-70B and LLaMa-2-13B for clarity's sake.

The \textit{prompt} is what is inputted to the LLM to generate its response. 
The prompt consists of three different parts: \textit{system message}, \textit{user message}, and \textit{assistant message}.
The system message sets the \textit{context} of the interaction with the LLM, as per the OpenAI official documentation~\footnote{\url{https://help.openai.com/en/articles/7042661-chatgpt-api-transition-guide}}.
The system message is used for instructing the model about the input format of the data points and ensuring that the output of LLMs conforms to a specific format. 
This pattern is also adopted by many open-source LLM models, including the LLaMa-2 models we consider in this work. 
For example, both the original LLaMa-2 weights and fine-tuned version use all three of these message types.

\textit{Prompt engineering} refers to building a specific prompt that works the best for the task. 
Prompt engineering includes designing how to wrap the task into a prompt and how to break the task down into smaller tasks that the LLM can handle. 

\noindent\textbf{Zero-shot learning.} Zero-shot prompts further assume that language models do not need any examples but understand the concept from training data. So, in these prompts, no examples are provided, but the LLM is directly prompted to predict a preference. \citet{kojimaLargeLanguageModels2022} first showed some zero-shot predictive capabilities of LLMs.

\noindent\textbf{Few-shot learning.} Few-shot prompts for language models~\cite{brownLanguageModelsAre2020} indicate the scenario where no weights are updated but a few examples are given to the language model to provide context. So, in our task of predicting preferences expressed in text, examples include a text, the names of the alternatives, and the expressed preference. Then, the model will receive instructions to predict preference for a new given text and pair of alternatives. The number of examples provided is a hyperparameter of the  algorithm. The name few-shot comes from the general concept of few-shot learning in ML~\cite{vinyals2016matching}.


\begin{table*}[!hpt]
\begin{tabular}{|l|p{2.5cm}|p{2.5cm}|p{2.5cm}|p{2.5cm}|}
\hline
\textbf{Model}         & \multicolumn{1}{l|}{\textbf{Cost per Output Token ($\times 10^{-6}$)}} & \multicolumn{1}{l|}{\textbf{Cost per Input Token ($\times 10^{-6}$)}} & \textbf{Architecture Size} & \textbf{Pre-train Token Amount} \\ \hline
\textbf{LlaMa-2 70B}   & 0                                                                      & 0                                                                     & 70B                        & 2T                              \\ \hline
\textbf{GPT-3.5-Turbo} & 2                                                                      & 1.5                                                                   & 175B*                      & unknown                         \\ \hline
\textbf{GPT-4}         & 60                                                                     & 30                                                                    & unknown                    & unknown                         \\ \hline
\end{tabular}
\caption{
    Statistics of individual models. The cost is measured in U.S. dollars. \cite{brownLanguageModelsAre2020,touvronLLaMAOpenEfficient2023a}
    * This value is taken from GPT-3 and it is difficult to confirm whether GPT-3.5-Turbo contains the same amount of parameters.
}
\label{tab:llm-model-stats}
\end{table*}

\section{Experiments}
In this work, we seek to assess the potential and limitations of using LLMs for classification tasks. 
We use an example from the prompts used for College Confidential to illustrate the workflow.

All of our code is publicly available on Github.\footnote{\url{https://github.com/inwonakng/llm-preference/}}

\subsection{Prompt Structure} 
For the College Confidential dataset, the system message starts with the sentences that tell the LLMs that it will be given one comment and two colleges and that its job is to identify the preference between the two colleges in this comment.
For Compsent-19, which does not have a single domain, we modify the role to ask it to assume the role of an internet forum user.
Some output rules are added to the system message to make the output conform to a specific format.
For example, we provide rules such as \textit{You MUST respond with ``A is preferred over B'' if college A is preferred over college B}.
Once we have an output that conforms to the desired format, we can assess the accuracy of LLMs in classification tasks in an automatic manner.

When sending the prompt to the LLM, a conversation is represented as a list of tuples. The first element of the tuple is the user's input, and the second is the LLM's output.
The final response from the model is triggered by sending a list of those tuples, followed by the last user instruction, to which the model will respond.
This structure allows us to simulate the \textit{history} of the interaction between the user and the model before asking for its response.

During our experiments, we find that using clear instructions and capitalizing the instruction words can lead the LLM to have less inconsistent responses.
For example, instead of saying \textit{do ... if ...}, using -\textit{you MUST ... if ...} leads to a more consistent and rule-following output from the LLM.
We test two versions of prompts with this setting.
The initial version of the prompt is referred to as \textit{long}, which has detailed rules and context in instruction. 
We also test with a paraphrased version of this prompt, which we refer to as \textit{short}. 

We also experiment with different hyperparameters available for the LLMs.
In particular, we use different values for \textit{temperature} and \textit{top\_p} of both LLaMa-2 and GPT-4.
These parameters control how the best response is chosen by the LLM.
Temperature is responsible for how the model chooses each token.
Lower temperature values lead the model to choose the tokens with more likelihood. 
Top P is used to pick a set of tokens that follow a previously selected token.
Given a selected token, the following set of tokens is chosen such that the set is the minimum number of tokens whose probability exceeds the P value.
We find that \textit{temperature} $= 1$ and \textit{top\_p} $= 0.7$ work the best for the \textit{short} prompt and \textit{temperature} $= 0.7$ and \textit{top\_p} $= 0.1$ work the best for the \textit{long} prompt.
\\
\begin{adjustwidth}{0.05cm}{0.05cm}
\begin{quote}
\textbf{Example of \textit{short} prompt} \\
\fontsize{8}{10}\selectfont
You will be given two colleges A and B, and a comment. Your job is to identify the preference between the two given colleges in the comment. \\
The names of the two colleges and the comment are delimited with triple backticks. \\
Here are the rules: \\
You MUST NOT use the colleges' real names. \\
You MUST refer to the colleges as A or B.  \\
You MUST respond with \triplebackticks No preference\triplebackticks if there is no explicit preference in the comment. \\
You MUST respond with \triplebackticks A is preferred over B\triplebackticks if college A is preferred over college B. \\
You MUST respond with \triplebackticks B is preferred over A\triplebackticks if college B is preferred over college A. \\
You MUST respond with \triplebackticks Equal preference\triplebackticks if colleges A and B are equally preferred. \\
You MUST respond with \triplebackticks No preference\triplebackticks, \triplebackticks A is preferred over B\triplebackticks, \triplebackticks B is preferred over A\triplebackticks, or \triplebackticks Equal preference\triplebackticks. \\
College A: \triplebackticks\{alternative\_a\}\triplebackticks \\
College B: \triplebackticks\{alternative\_b\}\triplebackticks \\
Comment: \triplebackticks\{text\}\triplebackticks \\
\end{quote}
\end{adjustwidth}

\subsection{Few-shot Learning} 
In few-shot learning, correct examples of the task are added to the prompt to guide the LLM.
For each label, we select one data point with that label in the dataset as part of the few-shot example and exclude those points from the testing set. 
In order to ensure that the examples contain enough content to be helpful while not increasing the prompt length too much, we select the text with the minimum length, which has more than 100 words for each label. As for zero-shot learning, the examples are simply an empty set, i.e., no example will be used. 

An example of the interaction tuple that is added to the chat history is as the following:
\begin{adjustwidth}{0.05cm}{0.05cm}
\begin{quote}
User:\\
\triplebackticks\
Comment: I would prefer Stanford rather than UCB.
\triplebackticks\

\triplebackticks\
Option A: Stanford University
\triplebackticks\

\triplebackticks\
Option B: UCB
\triplebackticks\
\\
Assistant:\\
\triplebackticks\
A is preferred over B
\triplebackticks\
\end{quote}
\end{adjustwidth}

\begin{figure*}[!htp]
    \centering
    \includegraphics[width=.8\textwidth]{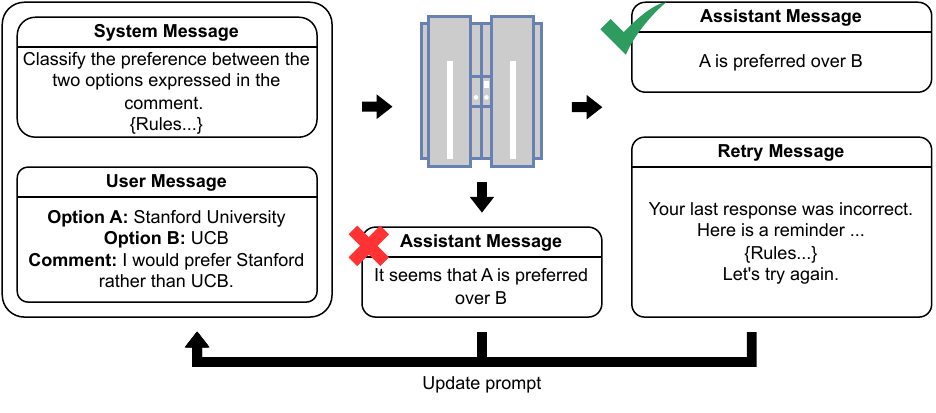}
    \caption{
        Illustration of the retry prompt process. The incorrect output is appended to the original prompt, followed by a retry message to remind the rules.
    }
    \label{fig:retry-prompt}
\end{figure*}

\subsection{Retry Prompting} 
Even if we specify that LLMs should produce specific content for each label, there is still a possibility that the LLMs may generate content that does not conform to the format. 
In some cases, we find that the model is correct in its classification, but we are not able to programmatically evaluate it due to the response being malformed. 
For example, instead of saying \textit{A is preferred over B}, the model can respond with \textit{... Therefore, I think A is preferred over B}.
To overcome cases such as this and allow for an automatic evaluation of the model's output, we use what we call a \textit{retry prompt}.
Instead of discarding the previously correct but malformed response, we re-build a prompt to continue the conversation and remind the model of the formatting rules again by adding another user message.
Using the tuple structure of the conversation history, we construct the retry prompt in a way that appears as the continuation of the task that was incorrectly formatted.
We append the tuple of the original task message and the model's incorrect response to the list of the conversation history. 
The retry message is then sent as the final user input, prompting the model to have access to its possibly correct but incorrectly formatted response to fix the format.
Using this technique, we are able to find some sets of rules with which the model's output was consistent for all the test inputs. 
Figure~\ref{fig:retry-prompt} shows an illustration of this process.

Specifically, a retry request is triggered when LLM produces a statement such as \textit{A is better than B in every way}, which deviates from the prescribed format. We use the same prompt rule as the previous short prompt rules in \{Rules...\}, and the retry prompt is as the following: \\

\begin{adjustwidth}{0.05cm}{0.05cm}
\begin{quote}
\textbf{Example of \textit{short} retry prompt} \\
\fontsize{8}{10}\selectfont
You have an incorrect format in your response. \\
Here is a reminder of the rules:\\
\{Rules...\} 
\end{quote}
\end{adjustwidth}


\section{Results}
\noindent In summary, we deploy the classification experiments in the following settings and analyze the results on LLaMa-2-70B, GPT-4, GPT-3.5-Turbo with long \& detailed prompt and short \& concise prompt with zero-shot and few-shot settings on College Confidential and Compsent-19.

\subsection{Prompt Engineering}
We experiment with various prompt techniques to convert the CPC task into a text format. 
Following best practices suggested by OpenAI\footnote{\url{https://help.openai.com/en/articles/6654000-best-practices-for-prompt-engineering-with-openai-api}} and DeepLearningAI\footnote{\url{https://www.deeplearning.ai/short-courses/chatgpt-prompt-engineering-for-developers/}}, we develop two sets of final prompts that are able used as input to the LLM to detect and classify preference in the text effectively.
For example, we set the role for the LLM to assume in the instruction (system) prompt and explain the rules in a clear bullet list. 
We also find that delimiters that are not commonly used in regular English text were best understood by the LLM. 
We use the triple-backtick \triplebackticks\  to wrap the comment and the two alternatives to note that they are separate from the instructions.
While GPT was also able to understand other delimiters, such as triple-hash (\#\#\#), this delimiter clashed with the template used by the instruct fine-tuned LLaMa-2 model used in the work.
Because of this reason, we use the triple-backtick delimiter for both models for consistency's sake.

When expressing the rules of the task, such as the formatting and the goal of the task, we find that a simple sentence structure in a commanding tone works best.
Interestingly, we also find that using a more conversational tone -- e.g. instead of ``Do XXX'', ``Let's do XXX'' -- helped the model follow the instructions more effectively.
For this reason, we express the rules themselves in a simple structure and capitalize the modal verbs -- e.g. must, must not -- and use a more conversational tone to end the prompt. 

\begin{table*}[!htp]
\begin{tabular}{|l|l|l|l|l|l|l|l|}
\hline
\textbf{Model} &
  \textbf{Prompt} &
  \textbf{Train Mode} &
  \textbf{F1 Micro} &
  \textbf{F1 Macro} &
  \textbf{F1{[}N/A{]}} &
  \textbf{F1{[}A \textgreater B{]}} &
  \textbf{F1{[}A \textless B{]}} \\ \hline
\multirow{4}{*}{\textbf{LLaMa-2 70B}} &
  \multirow{2}{*}{Short} &
  zero-shot &
  0.7287 &
  \textbf{0.6303} &
  0.8165 &
  \textbf{0.5445} &
  \textbf{0.5299} \\ \cline{3-8} 
& & few-shot  & \textbf{0.7381} & 0.6284 & \textbf{0.8264} & 0.5359 & 0.523 \\ \cline{2-8} 
& \multirow{2}{*}{Long}  & zero-shot & 0.7274 & 0.6111 & 0.8225 & 0.5265 & 0.4842 \\ \cline{3-8} 
& & few-shot  & 0.7247 & 0.5956 & 0.8186 & 0.4989 & 0.4692 \\ \hline
\multirow{4}{*}{\textbf{GPT-3.5-Turbo}} & \multirow{2}{*}{Short} & zero-shot & 0.6838          & 0.6165          & 0.7839          & 0.5402          & 0.5255          \\ \cline{3-8} 
& & few-shot  & \textbf{0.7054} & \textbf{0.6374} & 0.7919          & \textbf{0.5683} & \textbf{0.5522} \\ \cline{2-8} 
& \multirow{2}{*}{Long} & zero-shot & 0.6393 & 0.4636 & 0.7887          & 0.4160 & 0.1862 \\ \cline{3-8} 
& & few-shot  & 0.6841 & 0.4970 & \textbf{0.7987} & 0.3408 & 0.3516 \\ \hline
\multirow{4}{*}{\textbf{GPT-4}} & \multirow{2}{*}{Short} & zero-shot & 0.7213 & 0.6815 & 0.7945 & 0.6442 & 0.6058 \\ \cline{3-8} 
& & few-shot  & \textbf{0.7624} & \textbf{0.715}  & \textbf{0.8276} & \textbf{0.6755} & \textbf{0.6418} \\ \cline{2-8} 
& \multirow{2}{*}{Long} & zero-shot & 0.6879 & 0.6524 & 0.7682 & 0.6259 & 0.5629 \\ \cline{3-8} 
& & few-shot & 0.7304 & 0.6860 & 0.8050 & 0.6416 & 0.6113 \\ \hline
\end{tabular}
\caption{
    Comparison of LLM performance on College Confidential dataset.
}

\label{tab:colcon-performance}
\end{table*}

\begin{table*}
    \begin{tabular}{|l|l|l|l|l|l|}
    \hline
    \textbf{Model} &
    \textbf{F1 Micro} &
    \textbf{F1 Macro} &
    \textbf{F1{[}N/A{]}} &
    \textbf{F1{[}A \textgreater B{]}} &
    \textbf{F1{[}A \textless B{]}} \\ \hline
    \textbf{Best LLM} & \textbf{0.7624} & \textbf{0.7150} & \textbf{0.8276} & \textbf{0.6755} & \textbf{0.6418} \\ \hline
    \textbf{Best SotA} & 0.67* & 0.57\textsuperscript{\textdagger} & 0.79* & 0.60\textsuperscript{\textdagger} & 0.42\textsuperscript{\textdagger} \\ \hline
    \end{tabular}
    \caption{Comparison of best performance of LLM and SotA models on College Confidential dataset. * Results using SimCSEXGBoost as presented by \citet{mohsin2023cpc}. \textsuperscript{\textdagger} Results using MultiSentPref-20 as presented by \citet{mohsin2023cpc}.
    }
    \label{tab:colcon-sota-performance}
\end{table*}


\begin{table*}[!hpt]
\begin{tabular}{|l|l|l|l|l|l|l|l|}
\hline
\textbf{Model} &
  \textbf{Prompt} &
  \textbf{Train Mode} &
  \textbf{F1 Micro} &
  \textbf{F1 Macro} &
  \textbf{F1{[}N/A{]}} &
  \textbf{F1{[}A \textgreater B{]}} &
  \textbf{F1{[}A \textless B{]}} \\ \hline
\multirow{4}{*}{\textbf{LLaMa-2 70B}} &
  \multirow{2}{*}{Short} &
  zero-shot &
  0.7613 &
  0.6494 &
  0.8432 &
  0.6045 &
  0.5004 \\ \cline{3-8} 
& & few-shot     & \textbf{0.8524} & \textbf{0.7544} & \textbf{0.9063} & \textbf{0.7543} & 0.6027 \\ \cline{2-8} 
& \multirow{2}{*}{Long}  & zero-shot & 0.7969 & 0.6981 & 0.8672 & 0.6369 & 0.5902 \\ \cline{3-8} 
&& few-shot & 0.8521 & 0.7470 & \textbf{0.9091} & 0.7205 & \textbf{0.6114} \\ \hline
\multirow{4}{*}{\textbf{GPT-3.5-Turbo}} & \multirow{2}{*}{Short} & zero-shot & 0.5957 & 0.5473 & 0.6674 & 0.5302          & 0.4442 \\ \cline{3-8} 
& & few-shot & \textbf{0.8374} & \textbf{0.7212} & 0.8977 & \textbf{0.7048} & \textbf{0.5611} \\ \cline{2-8} 
& \multirow{2}{*}{Long} & zero-shot & 0.5347 & 0.4084 & 0.6410 & 0.4430 & 0.1413 \\ \cline{3-8} 
& & few-shot & \textbf{0.8374} & 0.6857 & \textbf{0.9030} & 0.6781 & 0.4759          \\ \hline
\multirow{4}{*}{\textbf{GPT-4}} & \multirow{2}{*}{Short} & zero-shot & 0.8149 & 0.7397 & 0.8739 & 0.7479          & 0.5974 \\ \cline{3-8} 
& & few-shot & 0.853 & 0.7792 & 0.9028 & \textbf{0.7939} & 0.6409 \\ \cline{2-8} 
& \multirow{2}{*}{Long} & zero-shot & 0.7839 & 0.7091 & 0.8493 & 0.7045          & 0.5736 \\ \cline{3-8} 
& & few-shot & \textbf{0.8580} & \textbf{0.7808} & \textbf{0.9083} & 0.7836 & \textbf{0.6506} \\ \hline

\end{tabular}
\caption{
    Comparison of LLM performance on Compsent-19 dataset.
}
\label{tab:compsent-performance}
\end{table*}

\begin{table*}
    \begin{tabular}{|l|l|l|l|l|l|}
    \hline
    \textbf{Model} &
  \textbf{F1 Micro} &
  \textbf{F1 Macro} &
  \textbf{F1{[}N/A{]}} &
  \textbf{F1{[}A \textgreater B{]}} &
  \textbf{F1{[}A \textless B{]}} \\ \hline
\textbf{Best LLM} & 0.8580 & \textbf{0.7808} & 0.9083 & \textbf{0.7939} & \textbf{0.6506} \\ \hline
\textbf{Best SotA} & \textbf{0.8743}\textsuperscript{$\ddagger$} & 0.7578\textsuperscript{$\ddagger$} & \textbf{0.9298}\textsuperscript{$\P$} & 0.7821\textsuperscript{$\ddagger$} & 0.5872\textsuperscript{$\ddagger$} \\ \hline
    \end{tabular}
    \caption{Comparison of best performance of LLM and SotA models on Compsent-19 dataset.
        \textsuperscript{$\ddagger$} Results using EDGAT\textsubscript{BERT}(8) as presented by \citet{ma2020edgat-compsent}. \textsuperscript{$\P$} Results using EDGAT\textsubscript{BERT}(9) as presented by \citet{ma2020edgat-compsent}.}
    \label{tab:compsent-sota-performance}
\end{table*}



\subsection{Classification Performance}
Tables~\ref{tab:compsent-performance} and ~\ref{tab:colcon-performance} show the results from our experiments with  LLaMa-2, GPT-3.5-Turbo, and GPT-4.

The best score for each dataset and model combination is highlighted in bold.
Because of the imbalance of labels in our dataset, we focus on both the Macro and Micro F1 scores, which calculate the unweighted and weighted averages of the individual F1 scores.

The results show that few-shot learning with the short prompt outputs the best performance in most cases. 
However, more detailed instructions may be helpful when the input text is shorter.
The model's performance on shorter text -- i.e. Compsent-19 -- in Table~\ref{tab:compsent-performance} shows that the zero-shot performance on LLaMa-2 70B with the long prompt can be better than that with the short prompt.
This suggests that the detailed instructions are helpful, but their effects are diminished when the other part of the prompt becomes lengthy.

It is also worth noting that the few-shot long prompt in Compsent-19 outperforms the few-shot short prompt more consistently with GPT-4. 
This suggests that GPT-4 may be more capable of handling complex/verbose instructions, especially when the input text itself is short.

While LLaMa-2 outperformed the previous state-of-the-art, the GPT-4 outperforms the LLaMa-2 models in general. However, LLaMa-2 is able to outperform GPT-3.5-Turbo.

In addition to the single-stage classification experiments, we consider another variety of prompting to handle long input. While Compsent-19 contains a single sentence per text, College Confidential dataset's text can be as long as multiple paragraphs. We design an experiment where the LLM first summarizes the input text and uses this summary to run the preference classification.
Specifically, the LLM is prompted to summarize the preference expressed in the text while ensuring the output contains the names of the two colleges.
This provides the necessary information for the subsequent preference classification task.

However, we find that the summary method does not perform as well as expected. When comparing using no summary versus using a summary, we see that no summary prompt outperforms the summary condition in the overall performance. When analyzing performance by text length, the summary approach only benefits texts longer than 400 words. Since posts in the College Confidential dataset tend to be shorter, the summary task may be unsuitable for this corpus. The poor performance of the summary may be due to the brevity of the source texts in College Confidential as well. 

\subsection{Output Consistency}
Throughout our experiments, we find that the LLM's responses tend to be inconsistent when faced with the same prompts. 
We observe that this happens often in the LLaMa-2 models and GPT-3.5-Turbo.
We note that GPT-4 was able to follow the rules much more consistently than LLaMa-2 -- while more than half of the tasks for LLaMa-2 had to be run with the retry prompt, only a handful of cases were needed for GPT-4.
For instance, the predicted label drastically changes if the same question is asked again, even though the response is correctly formatted.

We also find that the output from the few-shot is more likely to conform to the output format than zero-shot. It is likely because the examples show the correct outputs, and these examples can help LLM to do in-context learning and understand the output format rules.

The experiments on LLaMa-2 were deployed on the 4-bit quantized version of the 70B model.
We also deployed the experiments on the 13B model of LLaMa-2.
However, we observe that the LLaMa-2-13B model could not handle the College Confidential or Compsent-19 dataset. 
Specifically, the output consistency is hard to satisfy on LLaMa-2-13B, especially for zero-shot. 
On the other hand, LLaMa-2-13B tends to give fixed answers like “No preference” or “Equal preference” for most cases, which results in a bad performance.

\subsection{Detecting Preference}
As expected, the F1 scores for cases where preference is present are much lower than the \textit{N/A} cases, meaning the models struggled more with the classification task.
It is also worth noting that when the input text is a single sentence, as in the Compsent-19 dataset, the LLM tends to perform better in both detecting/classifying the preference. 
Overall, we again note that GPT-4 was able to outperform the LLaMa-2 models and GPT-3.5-Turbo in both detecting and classifying the preference.

\subsection{Comparison to Previous Work}
We compare our results against the state-of-the-art (SotA) results from previous works that leverage GNN-based models. 
\citet{ma2020edgat-compsent} present their results on the Compsent-19 dataset using the ED-GAT architecture, and \citet{mohsin2023cpc} present the results from MultiSentPref on the College Confidential dataset.
Tables~\ref{tab:colcon-sota-performance} and~\ref{tab:compsent-sota-performance} show the comparison between the best LLM and SotA performances. 

For College Confidential, we find that the LLM performance from GPT-4 outperforms the MultiSentPref's best performance by a significant margin.
Even LLaMa-2-70B can outperform every metric considered except for F1[A > B].

We also find that GPT-4 is able to outperform the best performance from ED-GAT in classifying the performance, albeit by a smaller margin than when compared to MultiSentPref.
It is worth noting here that ED-GAT performs better at detecting the lack of preference. 
This observation is also reflected in a higher F1 micro score as the dataset is unbalanced and contains approximately $2.7 \times$ more no preference rows as preference present rows.

It is interesting to note that MultiSentPref~\citep{mohsin2023cpc} is an extension of ED-GAT~\citep{ma2020edgat-compsent}, but it does not perform nearly as well in College Confidential. 
Notably, while Compsent-19 contains a single sentence per text, College Confidential's content can extend to multiple paragraphs; see Table~\ref{tab:stat_data}.
This disparity suggests that LLM's superiority over SotA likely arises from its capacity to manage extensive context, suggesting that it may be better at handling complex tasks compared to the previous graph-based SotA models.

Overall, we see an improvement over previous SotA models with both GPT-4 and LLaMa-2 in most metrics. GPT-4 consistently outperforms both MultiSentPref and ED-GAT in \textit{[preference classification} while LLaMa-2 is able to outperform MultiSentPref on College Confidential but falls short of ED-GAT on Compsent-19.
While ED-GAT outperforms in the preference detection task in Compsent-19, the difference is insignificant. 
The large improvement on the College Confidential dataset indicates that LLMs have significantly improved over previous SotA models in classifying longer context-length examples compared to smaller ones.

\section{Future Work}
Future extensions of this work can be branched into multiple directions.
The first direction would be to improve the LLM's performance by fine-tuning the model or improving the prompt. 
While fine-tuning the model will improve the performance, the lack of labeled datasets in the CPC domain poses a challenge.
The prompt engineering direction can include a more fine-grained approach to the problem, such as handling the two parts of CPC separately -- detecting the preference and classifying only if it exists -- or adding a summarization task before the classification to handle longer text better.
We considered a version of the summarization task in our work but were not able to find a set of prompts that led to a higher performance than the currently presented single-stage method.
Future works could explore more prompt engineering to find a set of prompts that allow the smaller models to handle large text more efficiently.

Another direction is to consider an ensemble learning approach of the LLMs. 
As seen in the comparison with the SotA models, models from previous work can outperform the current LLM approach. 
Thus, combining multiple previous models and LLMs in an ensemble could lead to better performance.
Another possibility is to consider the LLMs in an ensemble. 
The LLMs' predictions can be aggregated to form an ensemble output, resulting in better performance.

In another direction, we can generate new text datasets using an LLM to augment the original dataset to remedy the lack of comparative text datasets.
This newly augmented dataset could be used to train smaller models that usually need more data than currently available. Thus, a knowledge distillation process could be tried in which the preference predictive capabilities of LLMs can be distilled into smaller and more efficient models.

In a broader direction, we can explore the capabilities of LLMs in multi-agent scenarios such as group decision-making, deliberation, iterative decision-making, etc. We see much ongoing work in this domain, proposing to use LLMs for social choice~\cite{fish2023generative} or to facilitate multi-agent collaborations~\cite{gong2023mindagent,chen2023agentverse}. In this same vein, we can work to create an AI agent that assists in making better group decisions under uncertainty.

Finally, any application of an LLM for a specific task inherits the bias present in training the original LLMs. While LLaMa-2 was trained on open-source texts, this is not true for the GPT models, so we can not even be sure of the amount of bias in the models. Future work applying LLMs for preference learning and elicitation should further fine-tune the models to remove any possible biases since this is an even bigger issue when considering comparative texts. Also, while the LLM-based methods outperform state-of-the-art models in most cases, they still depend on black box methods. Thus, these methods should be applied cautiously in the real world. In the future, we will look into what steps we can take to make the preference prediction process more transparent, particularly focusing on getting explanations for the predictions.

\section{Conclusion}
In this work, we consider the task of predicting preferences expressed in text by using LLMs with an automated evaluation scheme.
Specifically, we experiment with four types of commercial and open-source LLMs -- OpenAI's GPT-4 and GPT-3.5-Turbo, Meta's LLaMa-2-70B and LLaMa-2-13B.
We also test the efficacy of different kinds of prompting methods to represent the task in a textual format and find two methods that are able to help the LLMs perform well in the task. 
Using these two prompts, we test the zero-shot and few-shot techniques to wrap the classification task into a conversational format that the LLM can handle.

Our results show that LLMs are able to outperform the previous SotA approaches in predicting preferences from text. 
We also find that using few-shot prompts by including examples from the dataset as a part of the prompt can further improve the LLM's performance.
While the smaller models are not able to handle a lengthy set of instructions and text as efficiently, we find that the larger model, such as  GPT-4, is able to handle both. 
The comparisons to previous SotA models show that the LLM is an effective replacement when handling longer texts. 
Some SotA approaches can still outperform our results using the LLMs when the input text is short. 
The observation that older methods can sometimes outperform LLMs suggests that we can use a sufficiently large enough LLM can be used in combination with other existing techniques for even better performances.
To summarize, our main findings are twofold: 1) LLM can be better than SotA models. 2) few-shot prompts are better than zero-shot ones.
  


\section*{Ethical Impacts}

Any application of LLMs for a specific task inherits the biases of the original LLM. We consider ethical concerns over this and discuss possible ways of alleviating some of those concerns in the Future Works section.



\bibliographystyle{ACM-Reference-Format} 
\bibliography{nlp,references}


\begin{thebibliography}{36}


\ifx \showCODEN    \undefined \def \showCODEN     #1{\unskip}     \fi
\ifx \showDOI      \undefined \def \showDOI       #1{#1}\fi
\ifx \showISBNx    \undefined \def \showISBNx     #1{\unskip}     \fi
\ifx \showISBNxiii \undefined \def \showISBNxiii  #1{\unskip}     \fi
\ifx \showISSN     \undefined \def \showISSN      #1{\unskip}     \fi
\ifx \showLCCN     \undefined \def \showLCCN      #1{\unskip}     \fi
\ifx \shownote     \undefined \def \shownote      #1{#1}          \fi
\ifx \showarticletitle \undefined \def \showarticletitle #1{#1}   \fi
\ifx \showURL      \undefined \def \showURL       {\relax}        \fi
\providecommand\bibfield[2]{#2}
\providecommand\bibinfo[2]{#2}
\providecommand\natexlab[1]{#1}
\providecommand\showeprint[2][]{arXiv:#2}

\bibitem[\protect\citeauthoryear{Boutilier}{Boutilier}{2002}]%
        {Boutilier02:POMDP}
\bibfield{author}{\bibinfo{person}{Craig Boutilier}.} \bibinfo{year}{2002}\natexlab{}.
\newblock \showarticletitle{A {POMDP} formulation of preference elicitation problems}. In \bibinfo{booktitle}{\emph{Proceedings of the National Conference on Artificial Intelligence (AAAI)}}. \bibinfo{address}{Edmonton, AB, Canada}, \bibinfo{pages}{239--246}.
\newblock


\bibitem[\protect\citeauthoryear{Brown, Mann, Ryder, Subbiah, Kaplan, Dhariwal, Neelakantan, Shyam, Sastry, Askell, Agarwal, {Herbert-Voss}, Krueger, Henighan, Child, Ramesh, Ziegler, Wu, Winter, Hesse, Chen, Sigler, Litwin, Gray, Chess, Clark, Berner, McCandlish, Radford, Sutskever, and Amodei}{Brown et~al\mbox{.}}{2020}]%
        {brownLanguageModelsAre2020}
\bibfield{author}{\bibinfo{person}{Tom Brown}, \bibinfo{person}{Benjamin Mann}, \bibinfo{person}{Nick Ryder}, \bibinfo{person}{Melanie Subbiah}, \bibinfo{person}{Jared~D Kaplan}, \bibinfo{person}{Prafulla Dhariwal}, \bibinfo{person}{Arvind Neelakantan}, \bibinfo{person}{Pranav Shyam}, \bibinfo{person}{Girish Sastry}, \bibinfo{person}{Amanda Askell}, \bibinfo{person}{Sandhini Agarwal}, \bibinfo{person}{Ariel {Herbert-Voss}}, \bibinfo{person}{Gretchen Krueger}, \bibinfo{person}{Tom Henighan}, \bibinfo{person}{Rewon Child}, \bibinfo{person}{Aditya Ramesh}, \bibinfo{person}{Daniel Ziegler}, \bibinfo{person}{Jeffrey Wu}, \bibinfo{person}{Clemens Winter}, \bibinfo{person}{Chris Hesse}, \bibinfo{person}{Mark Chen}, \bibinfo{person}{Eric Sigler}, \bibinfo{person}{Mateusz Litwin}, \bibinfo{person}{Scott Gray}, \bibinfo{person}{Benjamin Chess}, \bibinfo{person}{Jack Clark}, \bibinfo{person}{Christopher Berner}, \bibinfo{person}{Sam McCandlish}, \bibinfo{person}{Alec Radford}, \bibinfo{person}{Ilya Sutskever}, {and}
  \bibinfo{person}{Dario Amodei}.} \bibinfo{year}{2020}\natexlab{}.
\newblock \showarticletitle{Language {{Models}} Are {{Few-Shot Learners}}}. In \bibinfo{booktitle}{\emph{Advances in {{Neural Information Processing Systems}}}}, Vol.~\bibinfo{volume}{33}. \bibinfo{publisher}{{Curran Associates, Inc.}}, \bibinfo{pages}{1877--1901}.
\newblock


\bibitem[\protect\citeauthoryear{Chen, Su, Zuo, Yang, Yuan, Qian, Chan, Qin, Lu, Xie, et~al\mbox{.}}{Chen et~al\mbox{.}}{2023}]%
        {chen2023agentverse}
\bibfield{author}{\bibinfo{person}{Weize Chen}, \bibinfo{person}{Yusheng Su}, \bibinfo{person}{Jingwei Zuo}, \bibinfo{person}{Cheng Yang}, \bibinfo{person}{Chenfei Yuan}, \bibinfo{person}{Chen Qian}, \bibinfo{person}{Chi-Min Chan}, \bibinfo{person}{Yujia Qin}, \bibinfo{person}{Yaxi Lu}, \bibinfo{person}{Ruobing Xie}, {et~al\mbox{.}}} \bibinfo{year}{2023}\natexlab{}.
\newblock \showarticletitle{Agentverse: Facilitating multi-agent collaboration and exploring emergent behaviors in agents}.
\newblock \bibinfo{journal}{\emph{arXiv preprint arXiv:2308.10848}} (\bibinfo{year}{2023}).
\newblock


\bibitem[\protect\citeauthoryear{Chowdhery, Narang, Devlin, Bosma, Mishra, Roberts, Barham, Chung, Sutton, Gehrmann, Schuh, Shi, Tsvyashchenko, Maynez, Rao, Barnes, Tay, Shazeer, Prabhakaran, Reif, Du, Hutchinson, Pope, Bradbury, Austin, Isard, {Gur-Ari}, Yin, Duke, Levskaya, Ghemawat, Dev, Michalewski, Garcia, Misra, Robinson, Fedus, Zhou, Ippolito, Luan, Lim, Zoph, Spiridonov, Sepassi, Dohan, Agrawal, Omernick, Dai, Pillai, Pellat, Lewkowycz, Moreira, Child, Polozov, Lee, Zhou, Wang, Saeta, Diaz, Firat, Catasta, Wei, {Meier-Hellstern}, Eck, Dean, Petrov, and Fiedel}{Chowdhery et~al\mbox{.}}{2022}]%
        {chowdheryPaLMScalingLanguage2022}
\bibfield{author}{\bibinfo{person}{Aakanksha Chowdhery}, \bibinfo{person}{Sharan Narang}, \bibinfo{person}{Jacob Devlin}, \bibinfo{person}{Maarten Bosma}, \bibinfo{person}{Gaurav Mishra}, \bibinfo{person}{Adam Roberts}, \bibinfo{person}{Paul Barham}, \bibinfo{person}{Hyung~Won Chung}, \bibinfo{person}{Charles Sutton}, \bibinfo{person}{Sebastian Gehrmann}, \bibinfo{person}{Parker Schuh}, \bibinfo{person}{Kensen Shi}, \bibinfo{person}{Sasha Tsvyashchenko}, \bibinfo{person}{Joshua Maynez}, \bibinfo{person}{Abhishek Rao}, \bibinfo{person}{Parker Barnes}, \bibinfo{person}{Yi Tay}, \bibinfo{person}{Noam Shazeer}, \bibinfo{person}{Vinodkumar Prabhakaran}, \bibinfo{person}{Emily Reif}, \bibinfo{person}{Nan Du}, \bibinfo{person}{Ben Hutchinson}, \bibinfo{person}{Reiner Pope}, \bibinfo{person}{James Bradbury}, \bibinfo{person}{Jacob Austin}, \bibinfo{person}{Michael Isard}, \bibinfo{person}{Guy {Gur-Ari}}, \bibinfo{person}{Pengcheng Yin}, \bibinfo{person}{Toju Duke}, \bibinfo{person}{Anselm Levskaya},
  \bibinfo{person}{Sanjay Ghemawat}, \bibinfo{person}{Sunipa Dev}, \bibinfo{person}{Henryk Michalewski}, \bibinfo{person}{Xavier Garcia}, \bibinfo{person}{Vedant Misra}, \bibinfo{person}{Kevin Robinson}, \bibinfo{person}{Liam Fedus}, \bibinfo{person}{Denny Zhou}, \bibinfo{person}{Daphne Ippolito}, \bibinfo{person}{David Luan}, \bibinfo{person}{Hyeontaek Lim}, \bibinfo{person}{Barret Zoph}, \bibinfo{person}{Alexander Spiridonov}, \bibinfo{person}{Ryan Sepassi}, \bibinfo{person}{David Dohan}, \bibinfo{person}{Shivani Agrawal}, \bibinfo{person}{Mark Omernick}, \bibinfo{person}{Andrew~M. Dai}, \bibinfo{person}{Thanumalayan~Sankaranarayana Pillai}, \bibinfo{person}{Marie Pellat}, \bibinfo{person}{Aitor Lewkowycz}, \bibinfo{person}{Erica Moreira}, \bibinfo{person}{Rewon Child}, \bibinfo{person}{Oleksandr Polozov}, \bibinfo{person}{Katherine Lee}, \bibinfo{person}{Zongwei Zhou}, \bibinfo{person}{Xuezhi Wang}, \bibinfo{person}{Brennan Saeta}, \bibinfo{person}{Mark Diaz}, \bibinfo{person}{Orhan Firat},
  \bibinfo{person}{Michele Catasta}, \bibinfo{person}{Jason Wei}, \bibinfo{person}{Kathy {Meier-Hellstern}}, \bibinfo{person}{Douglas Eck}, \bibinfo{person}{Jeff Dean}, \bibinfo{person}{Slav Petrov}, {and} \bibinfo{person}{Noah Fiedel}.} \bibinfo{year}{2022}\natexlab{}.
\newblock \bibinfo{title}{{{PaLM}}: {{Scaling Language Modeling}} with {{Pathways}}}.
\newblock
\newblock
\urldef\tempurl%
\url{https://doi.org/10.48550/arXiv.2204.02311}
\showDOI{\tempurl}
\showeprint[arxiv]{2204.02311}~[cs]


\bibitem[\protect\citeauthoryear{Conitzer and Sandholm}{Conitzer and Sandholm}{2002}]%
        {Conitzer02:Elicitation}
\bibfield{author}{\bibinfo{person}{Vincent Conitzer} {and} \bibinfo{person}{Tuomas Sandholm}.} \bibinfo{year}{2002}\natexlab{}.
\newblock \showarticletitle{Vote Elicitation: Complexity and Strategy-Proofness}. In \bibinfo{booktitle}{\emph{Proceedings of the National Conference on Artificial Intelligence (AAAI)}}. \bibinfo{address}{Edmonton, AB, Canada}, \bibinfo{pages}{392--397}.
\newblock


\bibitem[\protect\citeauthoryear{Conneau, Kiela, Schwenk, Barrault, and Bordes}{Conneau et~al\mbox{.}}{2017}]%
        {conneau-EtAl:2017:EMNLP2017}
\bibfield{author}{\bibinfo{person}{Alexis Conneau}, \bibinfo{person}{Douwe Kiela}, \bibinfo{person}{Holger Schwenk}, \bibinfo{person}{Lo\"{i}c Barrault}, {and} \bibinfo{person}{Antoine Bordes}.} \bibinfo{year}{2017}\natexlab{}.
\newblock \showarticletitle{Supervised Learning of Universal Sentence Representations from Natural Language Inference Data}. In \bibinfo{booktitle}{\emph{Proceedings of EMNLP 2017}}. \bibinfo{publisher}{Association for Computational Linguistics}, \bibinfo{address}{Copenhagen, Denmark}, \bibinfo{pages}{670--680}.
\newblock
\urldef\tempurl%
\url{https://www.aclweb.org/anthology/D17-1070}
\showURL{%
\tempurl}


\bibitem[\protect\citeauthoryear{Devlin, Chang, Lee, and Toutanova}{Devlin et~al\mbox{.}}{2019}]%
        {devlin2019bert}
\bibfield{author}{\bibinfo{person}{Jacob Devlin}, \bibinfo{person}{Ming-Wei Chang}, \bibinfo{person}{Kenton Lee}, {and} \bibinfo{person}{Kristina Toutanova}.} \bibinfo{year}{2019}\natexlab{}.
\newblock \bibinfo{title}{BERT: Pre-training of Deep Bidirectional Transformers for Language Understanding}.
\newblock
\newblock
\showeprint[arxiv]{1810.04805}~[cs.CL]


\bibitem[\protect\citeauthoryear{Fish, G{\"o}lz, Parkes, Procaccia, Rusak, Shapira, and W{\"u}thrich}{Fish et~al\mbox{.}}{2023}]%
        {fish2023generative}
\bibfield{author}{\bibinfo{person}{Sara Fish}, \bibinfo{person}{Paul G{\"o}lz}, \bibinfo{person}{David~C Parkes}, \bibinfo{person}{Ariel~D Procaccia}, \bibinfo{person}{Gili Rusak}, \bibinfo{person}{Itai Shapira}, {and} \bibinfo{person}{Manuel W{\"u}thrich}.} \bibinfo{year}{2023}\natexlab{}.
\newblock \showarticletitle{Generative Social Choice}.
\newblock \bibinfo{journal}{\emph{arXiv preprint arXiv:2309.01291}} (\bibinfo{year}{2023}).
\newblock


\bibitem[\protect\citeauthoryear{Gao, Yao, and Chen}{Gao et~al\mbox{.}}{2021}]%
        {gao2021simcse}
\bibfield{author}{\bibinfo{person}{Tianyu Gao}, \bibinfo{person}{Xingcheng Yao}, {and} \bibinfo{person}{Danqi Chen}.} \bibinfo{year}{2021}\natexlab{}.
\newblock \showarticletitle{SimCSE: Simple Contrastive Learning of Sentence Embeddings}. In \bibinfo{booktitle}{\emph{Proceedings of EMNLP 2021}}.
\newblock


\bibitem[\protect\citeauthoryear{Gong, Huang, Ma, Vo, Durante, Noda, Zheng, Zhu, Terzopoulos, Fei-Fei, et~al\mbox{.}}{Gong et~al\mbox{.}}{2023}]%
        {gong2023mindagent}
\bibfield{author}{\bibinfo{person}{Ran Gong}, \bibinfo{person}{Qiuyuan Huang}, \bibinfo{person}{Xiaojian Ma}, \bibinfo{person}{Hoi Vo}, \bibinfo{person}{Zane Durante}, \bibinfo{person}{Yusuke Noda}, \bibinfo{person}{Zilong Zheng}, \bibinfo{person}{Song-Chun Zhu}, \bibinfo{person}{Demetri Terzopoulos}, \bibinfo{person}{Li Fei-Fei}, {et~al\mbox{.}}} \bibinfo{year}{2023}\natexlab{}.
\newblock \showarticletitle{MindAgent: Emergent Gaming Interaction}.
\newblock \bibinfo{journal}{\emph{arXiv preprint arXiv:2309.09971}} (\bibinfo{year}{2023}).
\newblock


\bibitem[\protect\citeauthoryear{Haque, Garg, Guo, and Singh}{Haque et~al\mbox{.}}{2022}]%
        {haque2022pixie}
\bibfield{author}{\bibinfo{person}{Amanul Haque}, \bibinfo{person}{Vaibhav Garg}, \bibinfo{person}{Hui Guo}, {and} \bibinfo{person}{Munindar~P Singh}.} \bibinfo{year}{2022}\natexlab{}.
\newblock \showarticletitle{Pixie: Preference in Implicit and Explicit Comparisons}. In \bibinfo{booktitle}{\emph{Proceedings of the 60th Annual Meeting of the Association for Computational Linguistics (Volume 2: Short Papers)}}. \bibinfo{pages}{106--112}.
\newblock


\bibitem[\protect\citeauthoryear{Hegselmann, Buendia, Lang, Agrawal, Jiang, and Sontag}{Hegselmann et~al\mbox{.}}{2023}]%
        {hegselmannTabLLMFewshotClassification2023}
\bibfield{author}{\bibinfo{person}{Stefan Hegselmann}, \bibinfo{person}{Alejandro Buendia}, \bibinfo{person}{Hunter Lang}, \bibinfo{person}{Monica Agrawal}, \bibinfo{person}{Xiaoyi Jiang}, {and} \bibinfo{person}{David Sontag}.} \bibinfo{year}{2023}\natexlab{}.
\newblock \showarticletitle{{{TabLLM}}: {{Few-shot Classification}} of {{Tabular Data}} with {{Large Language Models}}}. In \bibinfo{booktitle}{\emph{Proceedings of {{The}} 26th {{International Conference}} on {{Artificial Intelligence}} and {{Statistics}}}}. \bibinfo{publisher}{{PMLR}}, \bibinfo{pages}{5549--5581}.
\newblock
\showISSN{2640-3498}


\bibitem[\protect\citeauthoryear{Huang and Carley}{Huang and Carley}{2019}]%
        {huangSyntaxAwareAspectLevel2019}
\bibfield{author}{\bibinfo{person}{Binxuan Huang} {and} \bibinfo{person}{Kathleen Carley}.} \bibinfo{year}{2019}\natexlab{}.
\newblock \showarticletitle{Syntax-{{Aware Aspect Level Sentiment Classification}} with {{Graph Attention Networks}}}. In \bibinfo{booktitle}{\emph{Proceedings of the 2019 {{Conference}} on {{Empirical Methods}} in {{Natural Language Processing}} and the 9th {{International Joint Conference}} on {{Natural Language Processing}} ({{EMNLP-IJCNLP}})}}. \bibinfo{publisher}{{Association for Computational Linguistics}}, \bibinfo{address}{{Hong Kong, China}}, \bibinfo{pages}{5469--5477}.
\newblock
\urldef\tempurl%
\url{https://doi.org/10.18653/v1/D19-1549}
\showDOI{\tempurl}


\bibitem[\protect\citeauthoryear{Kojima, Gu, Reid, Matsuo, and Iwasawa}{Kojima et~al\mbox{.}}{2022}]%
        {kojimaLargeLanguageModels2022}
\bibfield{author}{\bibinfo{person}{Takeshi Kojima}, \bibinfo{person}{Shixiang~(Shane) Gu}, \bibinfo{person}{Machel Reid}, \bibinfo{person}{Yutaka Matsuo}, {and} \bibinfo{person}{Yusuke Iwasawa}.} \bibinfo{year}{2022}\natexlab{}.
\newblock \showarticletitle{Large {{Language Models}} Are {{Zero-Shot Reasoners}}}.
\newblock \bibinfo{journal}{\emph{Advances in Neural Information Processing Systems}}  \bibinfo{volume}{35} (\bibinfo{date}{Dec.} \bibinfo{year}{2022}), \bibinfo{pages}{22199--22213}.
\newblock


\bibitem[\protect\citeauthoryear{Li, Qin, Liu, and Wang}{Li et~al\mbox{.}}{2021}]%
        {li2021powering}
\bibfield{author}{\bibinfo{person}{Zeyu Li}, \bibinfo{person}{Yilong Qin}, \bibinfo{person}{Zihan Liu}, {and} \bibinfo{person}{Wei Wang}.} \bibinfo{year}{2021}\natexlab{}.
\newblock \showarticletitle{Powering Comparative Classification with Sentiment Analysis via Domain Adaptive Knowledge Transfer}. In \bibinfo{booktitle}{\emph{Proceedings of the EMNLP 2021}}.
\newblock


\bibitem[\protect\citeauthoryear{Liu, Ott, Goyal, Du, Joshi, Chen, Levy, Lewis, Zettlemoyer, and Stoyanov}{Liu et~al\mbox{.}}{2019}]%
        {liu2019roberta}
\bibfield{author}{\bibinfo{person}{Yinhan Liu}, \bibinfo{person}{Myle Ott}, \bibinfo{person}{Naman Goyal}, \bibinfo{person}{Jingfei Du}, \bibinfo{person}{Mandar Joshi}, \bibinfo{person}{Danqi Chen}, \bibinfo{person}{Omer Levy}, \bibinfo{person}{Mike Lewis}, \bibinfo{person}{Luke Zettlemoyer}, {and} \bibinfo{person}{Veselin Stoyanov}.} \bibinfo{year}{2019}\natexlab{}.
\newblock \showarticletitle{Roberta: A robustly optimized bert pretraining approach}.
\newblock \bibinfo{journal}{\emph{arXiv preprint arXiv:1907.11692}} (\bibinfo{year}{2019}).
\newblock


\bibitem[\protect\citeauthoryear{Ma, Mazumder, Wang, and Liu}{Ma et~al\mbox{.}}{2020}]%
        {ma2020edgat-compsent}
\bibfield{author}{\bibinfo{person}{Nianzu Ma}, \bibinfo{person}{Sahisnu Mazumder}, \bibinfo{person}{Hao Wang}, {and} \bibinfo{person}{Bing Liu}.} \bibinfo{year}{2020}\natexlab{}.
\newblock \showarticletitle{Entity-aware dependency-based deep graph attention network for comparative preference classification}. In \bibinfo{booktitle}{\emph{Proceedings of Annual Meeting of the Association for Computational Linguistics (ACL-2020)}}.
\newblock


\bibitem[\protect\citeauthoryear{Mandal, Shah, and Woodruff}{Mandal et~al\mbox{.}}{2020}]%
        {Mandal2020:Optimal}
\bibfield{author}{\bibinfo{person}{Debmalya Mandal}, \bibinfo{person}{Nisarg Shah}, {and} \bibinfo{person}{David~P. Woodruff}.} \bibinfo{year}{2020}\natexlab{}.
\newblock \showarticletitle{{Optimal Communication-Distortion Tradeoff in Voting}}. In \bibinfo{booktitle}{\emph{Proceedings of ACM EC}}.
\newblock


\bibitem[\protect\citeauthoryear{Mohsin, Kang, Chen, Shang, and Xia}{Mohsin et~al\mbox{.}}{2023}]%
        {mohsin2023cpc}
\bibfield{author}{\bibinfo{person}{Farhad Mohsin}, \bibinfo{person}{Inwon Kang}, \bibinfo{person}{Yuxuan Chen}, \bibinfo{person}{Jingbo Shang}, {and} \bibinfo{person}{Lirong Xia}.} \bibinfo{year}{2023}\natexlab{}.
\newblock \showarticletitle{Dependency and Coreference-boosted Multi-Sentence Preference model}. In \bibinfo{booktitle}{\emph{The 9th International Workshop on Deep Learning on Graphs: Method and Applications (DLG-AAAI-23)}}.
\newblock


\bibitem[\protect\citeauthoryear{Mohsin, Luo, Ma, Kang, Zhao, Liu, Vaish, and Xia}{Mohsin et~al\mbox{.}}{2021}]%
        {mohsin2021making}
\bibfield{author}{\bibinfo{person}{Farhad Mohsin}, \bibinfo{person}{Lei Luo}, \bibinfo{person}{Wufei Ma}, \bibinfo{person}{Inwon Kang}, \bibinfo{person}{Zhibing Zhao}, \bibinfo{person}{Ao Liu}, \bibinfo{person}{Rohit Vaish}, {and} \bibinfo{person}{Lirong Xia}.} \bibinfo{year}{2021}\natexlab{}.
\newblock \showarticletitle{Making group decisions from natural language-based preferences}. In \bibinfo{booktitle}{\emph{Proceedings of the 8th International Workshop on Computational Social Choice (COMSOC)}}.
\newblock


\bibitem[\protect\citeauthoryear{OpenAI}{OpenAI}{2022}]%
        {openaiIntroducingChatGPT2022}
\bibfield{author}{\bibinfo{person}{OpenAI}.} \bibinfo{year}{2022}\natexlab{}.
\newblock \bibinfo{title}{Introducing {{ChatGPT}}}.
\newblock \bibinfo{howpublished}{https://openai.com/blog/chatgpt}.
\newblock


\bibitem[\protect\citeauthoryear{OpenAI}{OpenAI}{2023}]%
        {openaiGPT4TechnicalReport2023}
\bibfield{author}{\bibinfo{person}{OpenAI}.} \bibinfo{year}{2023}\natexlab{}.
\newblock \bibinfo{title}{{{GPT-4 Technical Report}}}.
\newblock
\newblock
\urldef\tempurl%
\url{https://doi.org/10.48550/arXiv.2303.08774}
\showDOI{\tempurl}
\showeprint[arxiv]{2303.08774}~[cs]


\bibitem[\protect\citeauthoryear{Panchenko, Bondarenko, Franzek, Hagen, and Biemann}{Panchenko et~al\mbox{.}}{2019}]%
        {panchenko-etal-2019-categorizing}
\bibfield{author}{\bibinfo{person}{Alexander Panchenko}, \bibinfo{person}{Alexander Bondarenko}, \bibinfo{person}{Mirco Franzek}, \bibinfo{person}{Matthias Hagen}, {and} \bibinfo{person}{Chris Biemann}.} \bibinfo{year}{2019}\natexlab{}.
\newblock \showarticletitle{Categorizing Comparative Sentences}. In \bibinfo{booktitle}{\emph{Proceedings of the 6th Workshop on Argument Mining}}. \bibinfo{publisher}{Association for Computational Linguistics}, \bibinfo{address}{Florence, Italy}, \bibinfo{pages}{136--145}.
\newblock
\urldef\tempurl%
\url{https://doi.org/10.18653/v1/W19-4516}
\showDOI{\tempurl}


\bibitem[\protect\citeauthoryear{Panchenko, Ruppert, Faralli, Ponzetto, and Biemann}{Panchenko et~al\mbox{.}}{2018}]%
        {panchenko-etal-2018-building}
\bibfield{author}{\bibinfo{person}{Alexander Panchenko}, \bibinfo{person}{Eugen Ruppert}, \bibinfo{person}{Stefano Faralli}, \bibinfo{person}{Simone~P. Ponzetto}, {and} \bibinfo{person}{Chris Biemann}.} \bibinfo{year}{2018}\natexlab{}.
\newblock \showarticletitle{Building a Web-Scale -Parsed Corpus from {C}ommon{C}rawl}. In \bibinfo{booktitle}{\emph{Proceedings of the Eleventh International Conference on Language Resources and Evaluation ({LREC} 2018)}}. \bibinfo{publisher}{European Language Resources Association (ELRA)}, \bibinfo{address}{Miyazaki, Japan}.
\newblock
\urldef\tempurl%
\url{https://aclanthology.org/L18-1286}
\showURL{%
\tempurl}


\bibitem[\protect\citeauthoryear{Pennington, Socher, and Manning}{Pennington et~al\mbox{.}}{2014}]%
        {pennington2014glove}
\bibfield{author}{\bibinfo{person}{Jeffrey Pennington}, \bibinfo{person}{Richard Socher}, {and} \bibinfo{person}{Christopher~D Manning}.} \bibinfo{year}{2014}\natexlab{}.
\newblock \showarticletitle{Glove: Global vectors for word representation}. In \bibinfo{booktitle}{\emph{Proceedings of the EMNLP 2014}}. \bibinfo{pages}{1532--1543}.
\newblock


\bibitem[\protect\citeauthoryear{Radford and Narasimhan}{Radford and Narasimhan}{2018}]%
        {radfordImprovingLanguageUnderstanding2018}
\bibfield{author}{\bibinfo{person}{Alec Radford} {and} \bibinfo{person}{Karthik Narasimhan}.} \bibinfo{year}{2018}\natexlab{}.
\newblock \showarticletitle{Improving {{Language Understanding}} by {{Generative Pre-Training}}}.
\newblock


\bibitem[\protect\citeauthoryear{Schick and Sch{\"u}tze}{Schick and Sch{\"u}tze}{2021}]%
        {schickExploitingClozeQuestionsFewShot2021}
\bibfield{author}{\bibinfo{person}{Timo Schick} {and} \bibinfo{person}{Hinrich Sch{\"u}tze}.} \bibinfo{year}{2021}\natexlab{}.
\newblock \showarticletitle{Exploiting {{Cloze-Questions}} for {{Few-Shot Text Classification}} and {{Natural Language Inference}}}. In \bibinfo{booktitle}{\emph{Proceedings of the 16th {{Conference}} of the {{European Chapter}} of the {{Association}} for {{Computational Linguistics}}: {{Main Volume}}}}. \bibinfo{publisher}{{Association for Computational Linguistics}}, \bibinfo{address}{{Online}}, \bibinfo{pages}{255--269}.
\newblock
\urldef\tempurl%
\url{https://doi.org/10.18653/v1/2021.eacl-main.20}
\showDOI{\tempurl}


\bibitem[\protect\citeauthoryear{Touvron, Lavril, Izacard, Martinet, Lachaux, Lacroix, Rozi{\`e}re, Goyal, Hambro, Azhar, Rodriguez, Joulin, Grave, and Lample}{Touvron et~al\mbox{.}}{2023}]%
        {touvronLLaMAOpenEfficient2023a}
\bibfield{author}{\bibinfo{person}{Hugo Touvron}, \bibinfo{person}{Thibaut Lavril}, \bibinfo{person}{Gautier Izacard}, \bibinfo{person}{Xavier Martinet}, \bibinfo{person}{Marie-Anne Lachaux}, \bibinfo{person}{Timoth{\'e}e Lacroix}, \bibinfo{person}{Baptiste Rozi{\`e}re}, \bibinfo{person}{Naman Goyal}, \bibinfo{person}{Eric Hambro}, \bibinfo{person}{Faisal Azhar}, \bibinfo{person}{Aurelien Rodriguez}, \bibinfo{person}{Armand Joulin}, \bibinfo{person}{Edouard Grave}, {and} \bibinfo{person}{Guillaume Lample}.} \bibinfo{year}{2023}\natexlab{}.
\newblock \bibinfo{title}{{{LLaMA}}: {{Open}} and {{Efficient Foundation Language Models}}}.
\newblock
\newblock
\urldef\tempurl%
\url{https://doi.org/10.48550/arXiv.2302.13971}
\showDOI{\tempurl}
\showeprint[arxiv]{2302.13971}~[cs]


\bibitem[\protect\citeauthoryear{Vaswani, Shazeer, Parmar, Uszkoreit, Jones, Gomez, Kaiser, and Polosukhin}{Vaswani et~al\mbox{.}}{2017}]%
        {vaswani2017attention}
\bibfield{author}{\bibinfo{person}{Ashish Vaswani}, \bibinfo{person}{Noam Shazeer}, \bibinfo{person}{Niki Parmar}, \bibinfo{person}{Jakob Uszkoreit}, \bibinfo{person}{Llion Jones}, \bibinfo{person}{Aidan~N Gomez}, \bibinfo{person}{{\L}ukasz Kaiser}, {and} \bibinfo{person}{Illia Polosukhin}.} \bibinfo{year}{2017}\natexlab{}.
\newblock \showarticletitle{Attention is all you need}.
\newblock \bibinfo{journal}{\emph{Proceedings of NeurIPS 2017}} (\bibinfo{year}{2017}).
\newblock


\bibitem[\protect\citeauthoryear{Veličković, Cucurull, Casanova, Romero, Liò, and Bengio}{Veličković et~al\mbox{.}}{2018}]%
        {velickovic2018graph}
\bibfield{author}{\bibinfo{person}{Petar Veličković}, \bibinfo{person}{Guillem Cucurull}, \bibinfo{person}{Arantxa Casanova}, \bibinfo{person}{Adriana Romero}, \bibinfo{person}{Pietro Liò}, {and} \bibinfo{person}{Yoshua Bengio}.} \bibinfo{year}{2018}\natexlab{}.
\newblock \showarticletitle{Graph Attention Networks}. In \bibinfo{booktitle}{\emph{International Conference on Learning Representations}}.
\newblock


\bibitem[\protect\citeauthoryear{Vinyals, Blundell, Lillicrap, Wierstra, et~al\mbox{.}}{Vinyals et~al\mbox{.}}{2016}]%
        {vinyals2016matching}
\bibfield{author}{\bibinfo{person}{Oriol Vinyals}, \bibinfo{person}{Charles Blundell}, \bibinfo{person}{Timothy Lillicrap}, \bibinfo{person}{Daan Wierstra}, {et~al\mbox{.}}} \bibinfo{year}{2016}\natexlab{}.
\newblock \showarticletitle{Matching networks for one shot learning}.
\newblock \bibinfo{journal}{\emph{Advances in neural information processing systems}}  \bibinfo{volume}{29} (\bibinfo{year}{2016}).
\newblock


\bibitem[\protect\citeauthoryear{Wei, Bosma, Zhao, Guu, Yu, Lester, Du, Dai, and Le}{Wei et~al\mbox{.}}{2022a}]%
        {weiFINETUNEDLANGUAGEMODELS2022a}
\bibfield{author}{\bibinfo{person}{Jason Wei}, \bibinfo{person}{Maarten Bosma}, \bibinfo{person}{Vincent~Y Zhao}, \bibinfo{person}{Kelvin Guu}, \bibinfo{person}{Adams~Wei Yu}, \bibinfo{person}{Brian Lester}, \bibinfo{person}{Nan Du}, \bibinfo{person}{Andrew~M Dai}, {and} \bibinfo{person}{Quoc~V Le}.} \bibinfo{year}{2022}\natexlab{a}.
\newblock \showarticletitle{{{FINETUNED LANGUAGE MODELS ARE ZERO-SHOT LEARNERS}}}.
\newblock \bibinfo{journal}{\emph{International Conference on Learning Representations}} (\bibinfo{year}{2022}).
\newblock


\bibitem[\protect\citeauthoryear{Wei, Wang, Schuurmans, Bosma, Ichter, Xia, Chi, Le, and Zhou}{Wei et~al\mbox{.}}{2022b}]%
        {weiChainofThoughtPromptingElicits2022}
\bibfield{author}{\bibinfo{person}{Jason Wei}, \bibinfo{person}{Xuezhi Wang}, \bibinfo{person}{Dale Schuurmans}, \bibinfo{person}{Maarten Bosma}, \bibinfo{person}{Brian Ichter}, \bibinfo{person}{Fei Xia}, \bibinfo{person}{Ed Chi}, \bibinfo{person}{Quoc~V. Le}, {and} \bibinfo{person}{Denny Zhou}.} \bibinfo{year}{2022}\natexlab{b}.
\newblock \showarticletitle{Chain-of-{{Thought Prompting Elicits Reasoning}} in {{Large Language Models}}}.
\newblock \bibinfo{journal}{\emph{Advances in Neural Information Processing Systems}}  \bibinfo{volume}{35} (\bibinfo{date}{Dec.} \bibinfo{year}{2022}), \bibinfo{pages}{24824--24837}.
\newblock


\bibitem[\protect\citeauthoryear{Xia}{Xia}{2019}]%
        {xia2019learning}
\bibfield{author}{\bibinfo{person}{Lirong Xia}.} \bibinfo{year}{2019}\natexlab{}.
\newblock \showarticletitle{Learning and decision-making from rank data}.
\newblock \bibinfo{journal}{\emph{Synthesis Lectures on Artificial Intelligence and Machine Learning}} \bibinfo{volume}{13}, \bibinfo{number}{1} (\bibinfo{year}{2019}), \bibinfo{pages}{1--159}.
\newblock


\bibitem[\protect\citeauthoryear{Xia}{Xia}{2022}]%
        {xia2022group}
\bibfield{author}{\bibinfo{person}{Lirong Xia}.} \bibinfo{year}{2022}\natexlab{}.
\newblock \showarticletitle{Group decision making under uncertain preferences: powered by AI, empowered by AI}.
\newblock \bibinfo{journal}{\emph{Annals of the New York Academy of Sciences}} \bibinfo{volume}{1511}, \bibinfo{number}{1} (\bibinfo{year}{2022}), \bibinfo{pages}{22--39}.
\newblock


\bibitem[\protect\citeauthoryear{Zhao, Li, Wang, Kephart, Mattei, Su, and Xia}{Zhao et~al\mbox{.}}{2018}]%
        {Zhao2018:A-Cost-Effective}
\bibfield{author}{\bibinfo{person}{Zhibing Zhao}, \bibinfo{person}{Haoming Li}, \bibinfo{person}{Junming Wang}, \bibinfo{person}{Jeffrey Kephart}, \bibinfo{person}{Nicholas Mattei}, \bibinfo{person}{Hui Su}, {and} \bibinfo{person}{Lirong Xia}.} \bibinfo{year}{2018}\natexlab{}.
\newblock \showarticletitle{{A Cost-Effective Framework for Preference Elicitation and Aggregation}}. In \bibinfo{booktitle}{\emph{Proceedings of Uncertainty in Artificial Intelligence}}.
\newblock


\end{thebibliography}


\newpage
\section{Appendix}

\subsection{Full prompt for \textit{long}}

\begin{adjustwidth}{0.05cm}{0.05cm}
\begin{quote}
\textbf{Instruction Message} \\
\fontsize{8}{10}\selectfont
  Pretend that you are a user on college confidential forums.\\
  Your job is to detect if there exists a preference between two options in a comment. \\
  If there exists a preference, you must detect what the preference is.\\
  If the author of the comment expresses an explicit preference, you must detect it.\\
  You will be given a comment and two alternatives for each task.
  The options will be denoted by \triplebackticks\ Option A:\triplebackticks\ and \triplebackticks\ Option B:\triplebackticks\ .\\
  The comment will be denoted by \triplebackticks\ Comment:\triplebackticks\ .\\
  
  Rules:
  - You MUST NOT respond with a summary of the comment.\\
  - You MUST NOT use the options' real names.\\
  - You MUST refer to the options as A or B. \\
  - You MUST respond with "No preference``` if there is no strict preference.\\
  - You MUST respond with \triplebackticks\ A is preferred over B\triplebackticks\ if option A is preferred over option B.\\
  - You MUST respond with \triplebackticks\ B is preferred over A\triplebackticks\ if option B is preferred over option A.\\
  - You MUST respond with \triplebackticks\ Equal preference\triplebackticks\ if options A and B are equally preferred.\\
  - You MUST respond using one of the four phrases above. 
\end{quote}
\end{adjustwidth}

\begin{adjustwidth}{0.05cm}{0.05cm}
\begin{quote}
\textbf{Retry Message} \\
\fontsize{8}{10}\selectfont
  Your response was incorrect. \\
  Let's try again.\\
  Here is a reminder of the rules:\\

  - You MUST ONLY report the preference in the comment.\\
  - You MUST respond only using one of the following phrases: \triplebackticks\ No preference\triplebackticks\ , \triplebackticks\ A is preferred over B\triplebackticks\ , \triplebackticks\ B is preferred over A\triplebackticks\ , \triplebackticks\ Equal preference\triplebackticks\ . Do not say anything else.\\
  - You MUST respond with \triplebackticks\ No preference\triplebackticks\ if there is no strict preference.\\
  - You MUST respond with \triplebackticks\ A is preferred over B\triplebackticks\ if option A is preferred over option B.\\
  - You MUST respond with \triplebackticks\ B is preferred over A\triplebackticks\ if option B is preferred over option A.\\
  - You MUST respond with \triplebackticks\ Equal preference\triplebackticks\ if options A and B are equally preferred.\\
  - You MUST NOT use the options's real names.
  - You MUST ONLY refer to the options as \triplebackticks\ A\triplebackticks\ or \triplebackticks\ B\triplebackticks\ .\\
  - You MUST NOT respond with any other details than the preference expressed in the comment.\\
  - You MUST NOT explain your reasoning behind the response. Only respond with the given phrase.\\
  - You MUST NOT use any punctuation in the response.\\

  Your previous response was not in any of the required responses.\\
  Try again and respond with a correct response to the previous comment. \\
  You MUST NOT reply the same response.\\
\end{quote}
\end{adjustwidth}

\subsection{Full prompt for \textit{short}}
\begin{adjustwidth}{0.05cm}{0.05cm}
\begin{quote}
\textbf{Instruction Message} \\
\fontsize{8}{10}\selectfont
  You will be given two colleges A and B, and a comment. Your job is to identify the preference between the two given colleges in the comment. \\
  The names of the two colleges and the comment are delimited with triple backticks.\\

  Here are the rules:
  You MUST NOT use the colleges' real names.\\
  You MUST refer to the colleges as A or B. \\
  You MUST respond with \triplebackticks\ No preference\triplebackticks\ if there is no explicit preference in the comment.\\
  You MUST respond with \triplebackticks\ A is preferred over B\triplebackticks\ if college A is preferred over college B.\\
  You MUST respond with \triplebackticks\ B is preferred over A\triplebackticks\ if college B is preferred over college A.\\
  You MUST respond with \triplebackticks\ Equal preference\triplebackticks\ if colleges A and B are equally preferred.\\
  You MUST respond with \triplebackticks\ No preference\triplebackticks\ , \triplebackticks\ A is preferred over B\triplebackticks\ , \triplebackticks\ B is preferred over A\triplebackticks\ , or \triplebackticks\ Equal preference\triplebackticks\ .
\end{quote}
\end{adjustwidth}

\begin{adjustwidth}{0.05cm}{0.05cm}
\begin{quote}
\textbf{Retry Message} \\
\fontsize{8}{10}\selectfont
  You have an incorrect format in your response. \\
  Here is a reminder of the rules:\\
  You MUST NOT use the colleges' real names.\\
  You MUST refer to the colleges as A or B. \\
  You MUST respond with \triplebackticks\ No preference\triplebackticks\ if there is no explicit preference in the comment.\\
  You MUST respond with \triplebackticks\ A is preferred over B\triplebackticks\ if college A is preferred over college B.\\
  You MUST respond with \triplebackticks\ B is preferred over A\triplebackticks\ if college B is preferred over college A.\\
  You MUST respond with \triplebackticks\ Equal preference\triplebackticks\ if colleges A and B are equally preferred.\\
  You MUST respond with \triplebackticks\ No preference\triplebackticks\ , \triplebackticks\ A is preferred over B\triplebackticks\ , \triplebackticks\ B is preferred over A\triplebackticks\ , or \triplebackticks\ Equal preference\triplebackticks\ .
\end{quote}
\end{adjustwidth}
\end{document}